# Pseudo Replay-based Class Continual Learning for Online New Category Anomaly Detection in Advanced Manufacturing

Yuxuan Li[1], Tianxin Xie[2], Chenang Liu[3], Zhangyue Shi[4*]

[1]School of Business, East China University of Science and Technology, Shanghai, China,
[2]Department of Agricultural Economics, Oklahoma State University, Stillwater, OK, United States,
[3]School of Industrial Engineering & Management, Oklahoma State University, Stillwater, OK United States,
[4]Independent Researcher, Stillwater, OK, United States,
*Corresponding author, zhangyue.shi@outlook.com

**Abstract**: The incorporation of advanced sensors and machine learning techniques has enabled modern manufacturing enterprises to perform data-driven classification-based anomaly detection based on the sensor data collected in manufacturing processes. However, one critical challenge is that newly presented defect category may manifest as the manufacturing process continues, resulting in monitoring performance deterioration of previously trained machine learning models. Hence, there is an increasing need for empowering machine learning models to learn continually. Among all continual learning methods, memory-based continual learning has the best performance but faces the constraints of data storage capacity. To address this issue, this paper develops a novel pseudo replay-based continual learning framework by integrating class incremental learning and oversampling-based data generation. Without storing all the data, the developed framework could generate high-quality data representing previous classes to train machine learning model incrementally when new category anomaly occurs. In addition, it could even enhance the monitoring performance since it also effectively improves the data quality. The effectiveness of the proposed framework is validated in three cases studies, which leverages supervised classification problem for anomaly detection. The experimental results show that the developed method is very promising in detecting novel anomaly while maintaining a good performance on the previous task and brings up more flexibility in model architecture.

**Keywords**: Advanced manufacturing, continual learning, data generation, anomaly detection, pseudo replay.

# 1 Introduction

## 1.1 Research Background

The incorporation of advanced sensing technologies and artificial intelligence becomes the driving-force for smart manufacturing system (Liu, et al. 2022). Extensive sets of sensory data encompass a wealth of valuable process-related information that can be harnessed for the attainment of diverse objectives through a multitude of methodologies (Y. Li, Shi, and Liu 2023; Z. Shi, Mandal, et al. 2022). Process monitoring is one of the most important tasks in advanced manufacturing system given its profound influence on both product quality and cost (J. Shi 2023). In particular, machine learning, especially deep learning-based supervised process monitoring methods gain more and more attention as a result of their superior performance. Different from unsupervised monitoring which establishes decision boundaries based on in-control data and generates alarms for anomalies, the general procedure of supervised machine learning based monitoring consists of three steps: labeled process data collection, model training, and deployment of model for real-time monitoring. Currently there are already a couple of works successfully demonstrating the great potential of these methods in various manufacturing processes such as additive manufacturing, injection molding, laser machining, and laser welding (Z. Shi, Mamun, et al. 2022; Tercan, Deibert, and Meisen 2022; McDonnell et al. 2021; Ma et al. 2022).

The aforementioned studies are predicated upon an assumption, namely, that the available data are adequate for model training. Nonetheless, this assumption does not invariably align with the realities encountered in practical manufacturing production pipelines. Indeed, in real-world manufacturing system, process conditions may undergo fluctuations at any time and data are collected in an incremental time manner. For instance, in the context of additive manufacturing, alteration in infill rates and geometrical defects, which represent two distinct process anomalies, can manifest at different stages (see Figure 1). Such different types of anomalies in manufacturing will reflect different patterns in the sensor signals, i.e., different distributions (C. Liu et al. 2019; Rao et al. 2015). To detect these novel anomalies, we need to retrain our model. However, if directly training existed model on new data, model's performance on previously learned tasks abruptly degrades (i.e., catastrophic forgetting) (Shin et al. 2017). One intuition to avoid catastrophic forgetting is to retrain the model each time a novel condition arises. As data size scales up, owing to constraints stemming from hardware limitations or corporate policies, the long-term retention of process data is often unfeasible (Tercan, Deibert, and Meisen 2022; Sun et al. 2023). However, the imperative of conducting model training with all available data remains. There is an urgent demand for a methodology capable of facilitating the efficient incremental training of machine learning models during manufacturing processes.

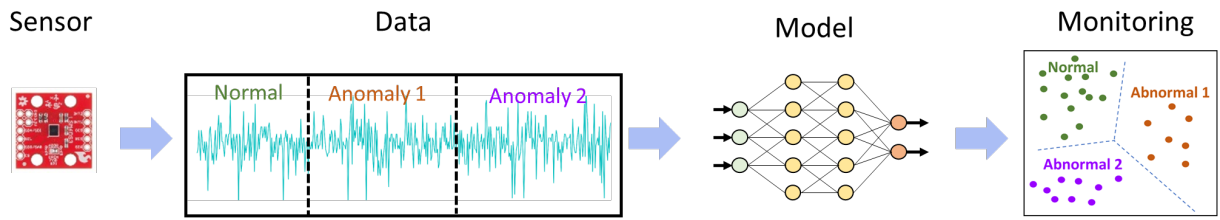

*Figure 1 Various process anomalies may manifest at distinct stages throughout the manufacturing process, necessitating machine learning models to discern novel defects in conjunction with existed ones.*

## 1.2 Research Objective

We have illustrated a specific challenge within the domain of advanced manufacturing, where the objective is to train a machine learning model to identify novel-class anomalies while maintaining a good performance on existing classes in manufacturing system. In this context, two distinct types of anomalies, namely infill rate alteration and geometry defects, occur sequentially during fabrication process (see more details in Sec 4). These anomalies, characterized by changes in alteration to design parameter (infill rate changes) and shape (geometry defects), represent different process conditions compared to the normal state. Essentially, this is a class-incremental problem, necessitating the machine learning model to progressively acquire the ability to discriminate between data from a growing number of classes.

In recent years, continual learning presents a promising avenue for mitigating this challenge. It constitutes a paradigm capable of sequentially training machine learning models, particularly neural networks, given a variety of tasks. The goal of continual learning is to avoid catastrophic forgetting while circumventing the need to access data from prior tasks. In general, continual learning methods consists of three categories: memory-based replay, parameter regularization, and functional regularization (van de Ven, Tuytelaars, and Tolias 2022). Notably, among these continual learning approaches, memory-replay methods, which stores all the existing data and retrains the model, have demonstrated superior efficacy in the context of class-incremental learning tasks (Hsu et al. 2018; van de Ven, Tuytelaars, and Tolias 2022). When new tasks arise, the impracticality of directly preserving all data collected within manufacturing systems and retraining model emerge stems from constraints associated with storage capacity.

In order to achieve enhanced monitoring model performance while alleviating the constraints posed by storage limitations, we develop a pseudo-replay continual learning framework. Rather than storing all the data directly, this framework employs a generator to learn data distribution for each individual task. When a new task emerges, the generator is leveraged to generate pseudo data derived from previous tasks. Subsequently, the model can be trained using the synthesized pseudo data together with the data from the new task. Compared to current class-incremental learning methods, the proposed method has the following advantages: (1) Performance and Hardware Efficiency: The pseudo-replay-based framework achieves a good balance between maintaining high model performance and alleviating hardware constraints. (2) Flexibility: Unlike regularization-based continual learning methods, the proposed replay-based framework offers enhanced flexibility. This flexibility allows for the use of different model architectures for previous and new models, facilitating the selection of the most appropriate model based on the characteristics of the data without being restricted by a fixed model structure. (3) Data Quality Enhancement: By employing a suitable generative model, the framework can generate more representative data, thus improving data quality and mitigating the impact of outliers, which could further improve the classification performance.

The rest of the structure of this paper is organized as follows. In Sec 2, a brief literature review is provided. The proposed research methodology framework is elaborated in Sec 3. To validate the effectiveness of proposed method, Sec 4 presents three real-world case studies. Eventually, conclusions are summarized, and future works are discussed in Sec 5.

# 2 Literature review and research background

## 2.1 Anomaly Detection in manufacturing system

Over the past decade, there has been a comprehensive exploration of anomaly detection (Haghnegahdar, Joshi, and Dahotre 2022; Arinez et al. 2020; Kandavalli et al. 2023). Various sensors fusion to different manufacturing processes has been a focal point, with corresponding signals leveraged for real-time process monitoring. Notably, vibration, acoustic emission, and temperature signals have found extensive applications in the detection of process anomalies, such as changes in layer thickness and nozzle clogging (Y. Li et al. 2021; Z. Shi, Li, and Liu 2022; J. Liu et al. 2018; Bastani, Rao, and Kong 2016). Furthermore, image data sourced from optical cameras and CT-imaging have been employed for the identification of defects such as porosity and part failure (Scime and Beuth 2018; M. Wang et al. 2022; Yao et al. 2018). Additionally, point cloud data, characterized by its richness in process information, has proven valuable for detecting shifts in the manufacturing process (Ye et al. 2021). It is noteworthy that video-based process monitoring is an emergent trend as well, exemplified by its application in detecting hot-spots in laser additive manufacturing and alterations in printing paths in fused filament fabrication (Yan et al. 2022; Al Mamun et al. 2022).

Regarding detection methodologies, a couple of analytical methods have been tailored to suit various manufacturing processes such as conventional statistical approaches encompass control charts (Al Mamun et al. 2022), Bayesian models (Bastani, Rao, and Kong 2016), and spatio-temporal regression (Yan et al. 2022). Moreover, domain-based methods, including fractal analysis (Yao et al. 2018), and dimension reduction methods such as principal component analysis (Z. Shi, Mamun, et al. 2022), linear discriminant analysis (J. Liu et al. 2018), and manifold learning (C. Liu et al. 2021), find widespread applications. Notably, machine learning-based methods have gained increasing attention in manufacturing, encompassing both supervised and unsupervised learning paradigms. Within the realm of supervised learning, some methodologies, including boosting (Z. Shi, Mamun, et al. 2022), random forest (Y. Li et al. 2021; Ye et al. 2021), and support vector machines (Gobert et al. 2018) are commonly employed. In addition, neural network-based approaches, including multi-layer perceptron (MLP) (J. Lee, Lee, and Kim 2020), convolutional neural network (CNN) (Z. Shi, Li, and Liu 2022; M. Wang et al. 2022; Z. Wang and Yao 2021), recurrent neural network (RNN) (G. Wang et al. 2019), and long short-term memory network (LSTM) (Z. Shi, Mamun, et al. 2022), constitute popular tools for effective *in-situ* process monitoring. On the unsupervised learning front, some techniques such as isolation forest (Y.-B. Wang et al. 2019), autoencoder (Z. Shi, Mamun, et al. 2022), clustering (Taheri et al. 2019), and one-class support vector machines (Z. Shi, Mamun, et al. 2022) are widely utilized.

Nevertheless, these methodologies follow the assumption that all pertinent data are accessible from the beginning, and the manufacturing process remains unaffected by the emergence of novel defects. However, such an assumption may not align with the realities encountered in actual production pipelines, where new process anomalies may manifest unpredictably. In such scenarios, the aforementioned methodologies prove inadequate for seamlessly incorporating new tasks while preserving superior performance on existing ones. Consequently, there arises a necessity for continual learning. In Sec 2.2, the state of the art of continual learning and its application in manufacturing is reviewed.

## 2.2 Continual learning and its application in manufacturing

For conventional machine learning models, they assume training data and test data are drawn from the same distribution, which is the basis for model's generalization capability for new unseen data. However, it may not hold true in real-world manufacturing systems where the training data for a learning task may only be available during certain time periods. Under this situation, a new model is needed when a novel class data arises. Continual learning is able to incrementally learn new skills without compromising those that were already learned (Kudithipudi et al. 2022). It is able to keep a good balance between the stability and flexibility of machine learning model as well.

The state-of-the-art continual learning methods consists of three categories: memory-based replay methods, parameter regularization methods, and functional regularization methods (van de Ven, Tuytelaars, and Tolias 2022). Memory-based methods either utilize a memory with fixed-size to store data from previously seen classes (Rebuffi et al. 2017), or utilize a pseudo-replay strategy to generate input data or latent features (Shin et al. 2017; Van de Ven, Siegelmann, and Tolias 2020). In terms of parameter regularization methods such as elastic weighted consolidation (EWC), memory-aware synapses (MAS), and synaptic intelligence, a regularization term is added to the loss function, which aims to penalizing important parameters for previously learnt tasks (Kirkpatrick et al. 2017; Aljundi et al. 2018; Zenke, Poole, and Ganguli 2017). When it comes to functional regularization methods, similar to parameter regularization, it adds a regularization term to the loss function which encourages the mapping between input and output not to change too much. Common methods include learn without forgetting (LwF) and Functional-Regularization of Memorable Past (FROMP) (Z. Li and Hoiem 2017; Pan et al. 2020). Among these continual learning methods, the memory-replay methods have the best performance for class-incremental learning (van de Ven, Tuytelaars, and Tolias 2022; Hsu et al. 2018).

In terms of application of continual learning in manufacturing, there are already some pioneering works. For parameter regularization based methods, EWC and Synaptic Intelligence have been applied to predict similar faults in turbofan engines using LSTM network (Maschler, Pham, and Weyrich 2021). In addition, Tercan *et al.* applied MAS for similar types of part deformation prediction in the injection modeling (Tercan, Deibert, and Meisen 2022). Li *et al.* froze several layers of the network to avoid significant weight change for rotary draw bending with mandrel process monitoring (J. Li et al. 2024). Speaking of functional regularization methods, only a few studies investigate the effectiveness of these methods in the turbofan fault prediction and defects inspection (Sun et al. 2023; Maschler, Pham, and Weyrich 2021). When it comes to replay-based methods, there are few applications which store data from previous tasks and retrain the model when new tasks come in. For example, Sen *et al.* evaluated effectiveness of replay-based continual learning strategy in three prediction tasks, tool wear, remaining useful lifetime, and process anomalies in CNC machining (Sen et al. 2023).

While continual learning shows great potential in anomaly detection, certain limitations still persist. In the context of parameter regularization and functional regularization methods, the model structure cannot change, as these methods necessitate the calculation of regularization loss based on existing neurons. This lack of flexibility precludes the adaptation of the model structure to varying data patterns. Furthermore, these methods exhibit optimal performance when trained on similar tasks. However, the diverse nature of anomalies in manufacturing systems may result in dissimilar patterns, compromising monitoring efficacy. For class-incremental tasks, it has been demonstrated that memory-based continual learning methods yield best performance (van de Ven, Tuytelaars, and Tolias 2022). As the volume of data increases, the extended retention of process data becomes impractical due to limitations imposed by hardware constraints.

## 2.3  Research Gaps

The research gap of class continual learning methods lies in several aspects. First, though memory-replay continual learning method based on data generation offers a viable solution by generating synthetic data highly similar to data from previous tasks and conserving storage memory, its exploration in manufacturing remains limited (S. Lee, Chang, and Baek 2021). Also, current replay-based continual learning methods struggle with hardware limitations to achieve a high-accuracy detection performance. In terms of regularization-based continual learning methods, they lack flexibility in model architecture. They are not allowed to use different model architectures for previous and new models, hindering the selection of the most appropriate model based on the characteristics of the data. In addition, for current continual learning methods, they do not consider improving model performance by enhancing the data quality. Therefore, in Sec 3, a pseudo memory-replay continual learning framework based on data generation is developed to address these gaps.

# 3 Research methodology

## 3.1 Research framework: replay-based continual learning (RCL)

In this section, a replay-based continual learning framework is developed based on pseudo data generation. The schematic representation of the research framework is presented in Figure 2, comprising two main components: training the generator sequentially and training the classifier with memory-replay. Initially, a generator is employed to learn the data distribution of each class following its manifestation in the manufacturing process. Various options for the generator exist, including methods such as the Synthetic Minority Oversampling Technique (SMOTE) (Chawla et al. 2002) and Generative Adversarial Network (GAN) (Goodfellow et al. 2014), which are proficient in learning underlying distributions and generating synthetic data. The generator, chosen based on its demonstrated effectiveness, generates pseudo data for each extant class without the necessity of storing the entire dataset. Subsequently, when a novel defect emerges, these trained generators can generate high-quality synthetic data representing pre-existing classes. This, in conjunction with newly collected data pertaining to the novel defect, facilitates the model update encompassing all classes. In this study, SMOTE is chosen as the generator for data augmentation for a better performance (See details in Sec 4). In Sec 3.2, basic information for SMOTE will be introduced, followed by the detailed paradigm in Sec 3.3.

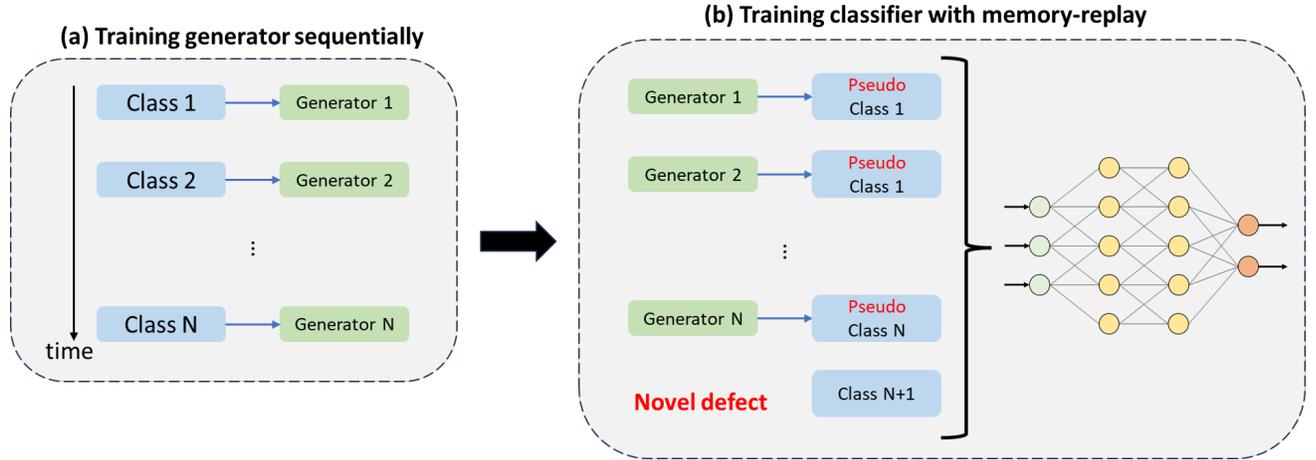

*Figure 2 (a) Sequentially training each generator involves training individual generators on their respective classes to acquire a comprehensive understanding of the distribution. (b) Upon the manifestation of a novel defect, the generators from previous tasks generate synthetic data of existing classes. These synthesized data, combined with the novel data, facilitate the training of an updated model with the ability to discriminate between prior and novel classes.*

## 3.2  SMOTE for pseudo class generation

In this section, the synthesis of high-quality artificial data representing pre-existing classes is facilitated through the application of SMOTE (Chawla et al. 2002). Originally designed to address data imbalanced issues, i.e., wherein the number of minority samples is significantly lower than that of majority samples, SMOTE operates by oversampling each minority class sample. The oversampling is achieved through synthesis samples along the line segments connecting less than or equal to $k$ nearest neighbors in the minority class, where $k$ serves as a tuning parameter. Illustrated in Figure 3, assuming $k$ is set to 3, for a given minority point $x_1$, three additional minority class nearest neighbors ($x_2$ to $x_4$) are selected. Subsequently, along each line segment connecting $x_1$ to another minority class nearest neighbor, a point is randomly chosen, yielding synthetic samples denoted as $a$, $b$, and $c$. Specifically, if the desired number of synthetic minority samples equals $k$ times the number of actual minority samples, all points $a$, $b$, and $c$ are produced as synthetic minority samples. Alternatively, if the required number of samples is not met, only a subset of the obtained points is generated. In this way, SMOTE facilitates the augmentation of minority class samples.

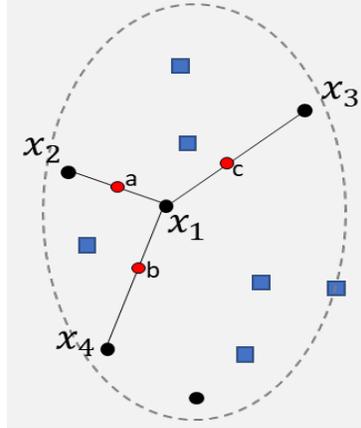

Figure 3 A demonstration of SMOTE algorithm. For actual sample $x_1$, three nearest neighbors $x_2$, $x_3$ and $x_4$ are selected and then three new synthetic points $a$, $b$ and $c$ are calculated.

It is imperative to note that minority data augmentation through SMOTE is conducted independently of the majority samples. Consequently, SMOTE can be conceptualized as a data augmentation technique designed specifically for an individual group, rather than being confined to either the minority or majority class. Given the presence of $N$ classes in this study, SMOTE can be employed autonomously for each class, functioning as generator 1 through generator $N$. Subsequently, pseudo classes 1 through pseudo class $N$ could be obtained in training the classifier with memory-replay mechanisms.

Algorithm 1 outlines the SMOTE algorithm applied in the context of replay-based continual learning. For each sample belonging to class $i$, the algorithm first acquires its $k$ nearest neighbors. Subsequently, a random point along the line segment connecting the given sample to each of its nearest neighbors is computed. The number of synthetic samples could be determined according to the requirements. From the resulting set of calculated points, if the number of generated samples needs to be equal to the number of original samples, only one point is then randomly selected to serve as the synthetic sample. Otherwise, several points will be selected. Then all the synthetic samples could be combined as pseudo class $i$ for next step, which is detailed discussed in Sec 3.3.

**Algorithm 1** SMOTE algorithm

---

**Input**: Samples in each class $\mathbf{X}_1,\ldots,\mathbf{X}_N$; Number of nearest neighbors $k$; Number of generated samples $S$
**For** class $i$ **from** 1 **to** $N$:
   **Step 1:** Denote number of samples in $\mathbf{X}_i$ as $M$
   **For** sample $\mathbf{X}_i^j$ **from** 1 **to** $M$:
     **Step 2:** Calculate $k$ nearest neighbors $x_1,\ldots, x_j$ to $\mathbf{X}_i^j$
     **For** $x_l$ **from** 1 **to** k:
       **Step 3:** Obtain one random point as $\mathbf{x}_{syn_{l_i}}^j$ along the line segment between $\mathbf{X}_i^j$ and $x_l$
     **Step 4:** Choose $[\frac{S}{N}]$ random points as $\mathbf{x}_{syn_i}^j$ among $\{\mathbf{x}_{syn_{1_i}}^j, \ldots, \mathbf{x}_{syn_{k_i}}^j\}$
   **Step 5:** Combine $\{\mathbf{x}_{syn_i}^1, \ldots, \mathbf{x}_{syn_i}^M\}$ as pseudo class $i$
**Output:** Pseudo class 1, **…,** Pseudo class $N$

---

## 3.3 Replay-based continual learning in manufacturing

Without loss of generality, this section elaborates the pseudo replay-based continual learning framework with two process anomalies (i.e., geometry defects and infill rate alteration) which manifest sequentially during manufacturing, as demonstrated in Figure 4. This could be further extended to process monitoring with more anomalies.

It is assumed that the product is fabricated normally at the early stage. Data from normal status could be collected and the first generator (i.e., generator 1) could be trained accordingly. Generator deploys SMOTE to learn the data pattern. When the first anomaly (i.e., infill rate change) in the study manifests, generator 2 could be trained to learn the pattern from data with altered infill rate. Meanwhile, a classifier could be trained with the real data that collected from infill rate change together with synthetic data representing normal status generated from generator 1 to distinguish between normal and infill rate alteration. Afterwards, when geometry defect manifests, generator 3 could be deployed to learn the underlying pattern. Pseudo normal and pseudo infill rate alteration data can be generated by generator 1 and 2, respectively. Then a new classifier can be trained using real geometry defect data and synthetic data from generators to differentiate normal status and these two process anomalies.

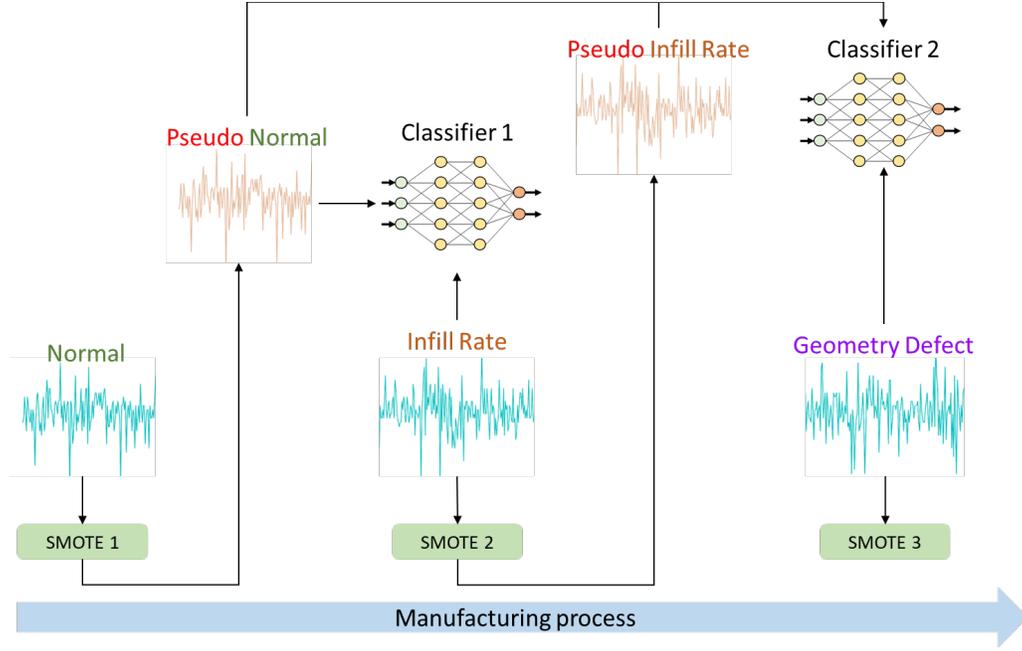

*Figure 4 Replay-based continual learning framework to detect infill rate alteration and geometry defect sequentially.*

---

**Algorithm 2** Replay-based Incremental Learning for Multiple Anomalies

---

**Input**: Samples in normal class $\mathbf{X}_0$ and different anomalies $\mathbf{X}_1,\ldots,\mathbf{X}_N$;
**Step 1:** Train generator $\mathbf{G}_0$ for normal class $\mathbf{X}_0$
**For** class $i$ **from** 1 **to** $N$:
   **Step 2:** Train generator $\mathbf{G}_i$ for anomaly $\mathbf{X}_i$
   **Step 3:** Generate pseudo data representing existing classes $\widetilde{\mathbf{X}}_0,\ldots,\widetilde{\mathbf{X}}_{i-1}$;
   **Step 4:** Train classifier $\mathbf{C}_i$ using $\{\widetilde{\mathbf{X}}_0,\ldots,\widetilde{\mathbf{X}}_{i-1},\mathbf{X}_i\}$
**Output:** Classifier $\mathbf{C}_i$

---

This framework, as delineated in Algorithm 2, could be systematically expanded to sequentially learn tasks involving multiple classes. The generator for the normal class, denoted as $\mathbf{G}_0$, is invariably the first to be trained. In the event of a process anomaly, a new generator, $\mathbf{G}_i$, is employed to grasp the underlying pattern of the anomaly. Subsequently, pseudo data from preceding classes are generated by the generators $\mathbf{G}_0$ through $\mathbf{G}_{i-1}$. These pseudo data, in conjunction with the new anomaly data, enable the training of a new classifier, $\mathbf{C}_i$, which is capable of distinguishing all previously observed process anomalies. In practical applications, the selection of an appropriate model, $C_i$, can be guided by domain knowledge. In the absence of such knowledge, it is advisable to evaluate a range of popular and powerful models, including tree-based methods, multilayer perceptrons (MLP), convolutional neural networks (CNN), long short-term memory networks (LSTM), and transformers. Subsequently, an appropriate model can be identified based on a comprehensive assessment of model performance and complexity.

The proposed replay-based continual learning (RCL) framework presents several noteworthy advantages. Firstly, among various types of continual learning methods, the replay-based approach demonstrates optimal performance for class-incremental tasks, aligning seamlessly with the practical requirements of manufacturing systems. Concurrently, the strategy of pseudo data generation addresses the constraints imposed by hardware limitations as data size scales because only few more representative data need to be stored. The data which is more representative means the data includes less outlier samples and can improve the classification performance. For instance, when generating samples based on original outlier samples, the linear combination could help to mitigate the influence of outlier samples. Sec. 4 will also validate the data quality improvement of the proposed method by integrating SMOTE.

Moreover, in contrast to other continual learning methods, the replay-based approach affords enhanced flexibility, allowing for disparate model architectures between the previous and new models according to divergent data characteristics. It is pertinent to highlight that the selection of SMOTE as the generator has the possibility of further improving data quality by producing more representative data compared to the original dataset and mitigating the impact of extreme data. This could potentially enhance monitoring performance when compared to models trained exclusively with original data. In the next section, three case studies in real-world application are conducted to validate the effectiveness of the proposed framework.

## 4   Real-world case study

To ascertain the efficacy of the proposed methodology, this section undertakes an evaluation of its performance within three different real-world datasets. The first two datasets are public datasets called Steel Plate Faults and Robot Execution Failure from UCI Machine Learning Repository (M Buscema 2010; Luis and Luis 1999), which is demonstrated in Sec 4.1 and Sec. 4.2, respectively. The third dataset is validated in Sec 4.3, which is a private dataset from the context of a practical additive manufacturing (AM) process, specifically focusing on fused filament fabrication (FFF).

## 4.1 A real-world case study based on steel plate faults

In this section, a public dataset of the Steel Plate Faults serves as an illustrative example for showcasing class-incremental tasks (M Buscema 2010). This multivariate dataset comprises 1941 samples and 27 independent variables, initially curated for classification purposes. Within the scope of this study, three distinct categories of steel plate faults—Z_Scratch, Pastry, and Bumps—are employed to formulate the class-incremental tasks. We assume that these three faults manifest sequentially, encompassing Task 1 and Task 2 as delineated below:

- **Task 1:** Classify both Z_scratch and Pastry faults from the collected data
- **Task 2:** When the third type of faults, i.e., Bumps, comes in, classify all three faults from the collected data

To demonstrate the efficacy of the proposed methodology, a comparative analysis is undertaken by implementing several benchmark approaches. Primarily, the classic incremental learning approach, Elastic Weight Consolidation (EWC) (Kirkpatrick et al. 2017), is employed. A Multi-Layer Perceptron (MLP) classifier with three fully connected layers is applied as the classifier. To be consistent, all the approaches in this paper utilize the same MLP as the base classifier. Besides, standardization is applied to all samples as well to ensure compatibility with the neural network structure. Subsequently, an approach referred to as fine-tuning, having the same MLP model structure is also considered as a benchmark method (Kirkpatrick et al. 2017). That is, the model is initially trained by the data from Z_scratch and Pastry. Afterwards, the model is trained by the normal data and the data from Bumps. Then the model is applied to accomplish Task 2. For a fair comparison, the baseline, representing classification without any incremental learning approach, i.e., training model with data from all three classes at the same time, is used as another benchmark. Each approach is trained and tested in an ensemble way with five base models for each task.

It is important to note that, the other generative models, such as generative adversarial networks and variational autoencoder may not be suitable applied in this work because the limited sample size is not able to satisfy the training requirements of these neural network-based approaches. To better illustrate that, t-distributed stochastic neighbor embedding (T-SNE) algorithm is applied in this case to visualize all the original data and synthetic data from RCL, conditional GAN (CGAN) and variational autoencoder (VAE). All the three approaches are trained by all the samples and generate the same number of samples as the original data. The 2-d T-SNE plot shown in Figure 5. Figure 5 (a) demonstrates the three groups in actual samples while Figure 5 (b) demonstrates the three groups in RCL-based samples compared with actual data. It is shown that they have a similar pattern for all three groups. Thus, it is proved that the proposed method is able to generate synthetic samples which are similar to actual samples. As for the T-SNE plot from CGAN-synthesis data (Figure 5 (c)) and VAE-synthesis data (Figure 5 (d)), the generated samples are completely different from the original samples, meaning that both CGAN and VAE approaches do not learn the data pattern very well.

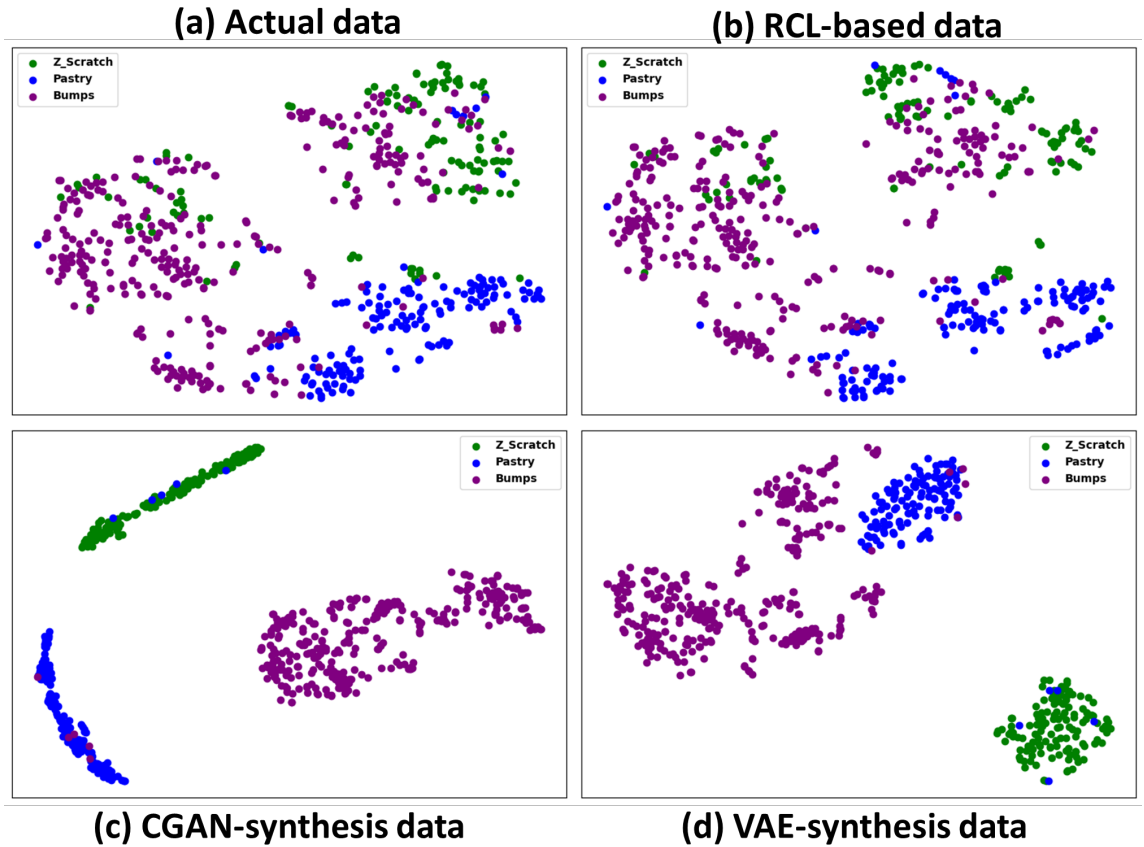

*Figure 5 The T-SNE plot of actual data (a), RCL-based data (b), CGAN-synthesis data (c) and VAE-synthesis data (d).*

Figure 6 illustrates the average precision, recall, and F-scores for all considered approaches pertaining to both Task 1 and Task 2. The detailed numerical values for these metrics are provided in TABLE 1, with the standard deviations enclosed in brackets. It is important to note that, given the commonality in the initial step across all approaches involving training the binary classifier, the metrics for Task 1 may exhibit identical values, specifically 0.962. Consequently, there remains an opportunity for enhancement in the realm of accurately identifying data corresponding to Task 1.

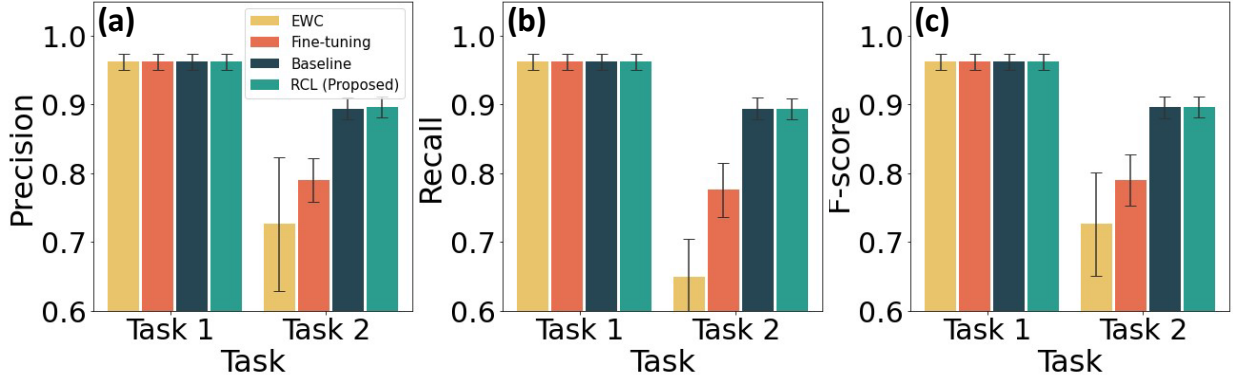

*Figure 6 The precisions (a), recalls (b) and F-scores (c) of different approaches on Task 1 and 2.*

*TABLE 1:* Average precisions, recalls and F-scores of Task 1 and 2

| Method | Task 1 | | | Task 2 | | |
|---|---|---|---|---|---|---|
| | Precision | Recall | F-score | Precision | Recall | F-score |
| EWC | 0.962 (0.012) | 0.962 (0.012) | 0.962 (0.012) | 0.726 (0.097) | 0.650 (0.055) | 0.610 (0.075) |
| Fine-tuning | | | | 0.790 (0.032) | 0.776 (0.039) | 0.774 (0.037) |
| Baseline | | | | 0.894 (0.016) | 0.894 (0.016) | 0.894 (0.016) |
| **RCL (Proposed)** | | | | **0.896 (0.015)** | **0.894 (0.015)** | **0.894 (0.015)** |

In the context of Task 2, the proposed methodology demonstrates superior precision, recall, and F-scores compared to all benchmark approaches, indicating heightened efficacy in the detection of both pre-existing and novel steel plate faults. It is noteworthy that, conventionally, the performance of continual learning-based approaches often fails to surpass the baseline as shown in TABLE 1. Although the F-score of the proposed method does not surpass that of the baseline, it remains equivalent to the baseline, while the metrics values for all other approaches register considerably lower. Furthermore, the standard deviation associated with the proposed method is markedly lower than that of other approaches and is comparable to the baseline, indicative of the proposed method's capacity to consistently generate more representative data. Specifically, the performance of CGAN and VAE approaches are also tested. However, the performance of CGAN and VAE models is completely worse than the proposed method. For instance, the average F-score of CGAN for Task 2 is about 0.45 and the average F-score of VAE for Task 2 is about 0.40. Comparing to them, the average F-score of the proposed method for Task 2 is about 0.89, which fully outperforms both the CGAN and VAE approaches.

In addition, the sample size requirement for SMOTE algorithm is also discussed in this section. To better illustrate the number of samples SMOTE needed, experiments regarding the relationship between the precision, recall, and F-score of the proposed method and the number of original samples are conducted. It is important to note that, the number of generated samples remains the same, i.e., 120 samples, under different number of original samples for generator training. The results are shown in Figure 7. The red solid lines are the metrics of the proposed method while the green dashed lines are the metrics of baseline. As the number of samples increases, the metrics will increase as well. Specifically, when the number of original training samples is about 20, the metrics of the proposed method are already higher than 0.8 and the differences between the proposed method and baseline are less than 10%. Hence, 20 samples to train SMOTE should be able to have convincing results. This outcome substantiates the effectiveness of the SMOTE-based memory-play continual learning framework since it could achieve comparable performance while alleviating the constraints of data storage. Afterwards, the other two case studies are conducted to further validate the effectiveness of the RCL.

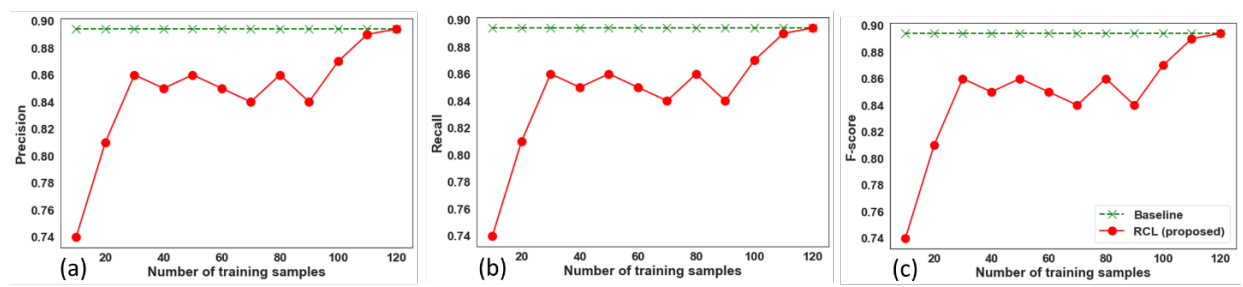

*Figure 7 The relationship between number of training samples and precision (a), recall (b) and F-score (c).*

## 4.2 A real-world case study based on robot execution failure

In this section, a public dataset of Robot Execution Failure is applied to demonstrate the effectiveness of the proposed method under a high-dimensional and high-number-of-class case (Luis and Luis 1999). The dataset includes 379 instances, 90 features from 7 classes. It contains force and torque measurements on a robot after failure detection. These 7 classes consist of one normal class and six different kinds of failure, including "Collision", "Obstruction", "Collision_in_part", "Bottom_collision", "Collision_in_tool" and "Bottom_obstruction". Afterwards, the approaches similar to Sec 4.1, including EWC, fine-tuning approach, and the proposed RCL approach are applied. The performance evaluation is conducted through the measurement of classification metrics, specifically precisions, recall and F-score as well. It is important to note that, though the dimension of the dataset is high, the sample size of different classes is extremely different. For instance, "Collision" has the largest number of samples, about 109 samples, while "Bottom_obstruction" has the smallest number of samples, about 23 samples. Hence, the neural network-based generative models are not tested in this case due to the insufficient data to train the models.

Since there are seven classes, the class-incremental tasks are structured as follows: In every task, a new type of failure will come in, and the model needs to identify all the types of failure. Hence, there will be 6 tasks, from Task 1 to Task 6. Specifically, the order of failure type is based on the number of samples, which means the type of failure with the highest number of samples will come in first. Thus, "Collision" class will be the first class to be classified with the normal class. Then the classes of "Obstruction", "Collision_in_part", "Bottom_collision", and "Collision_in_tool" are sent to the model gradually. Finally, the class of "Bottom_obstruction" comes in.

To show that the proposed method is also able to generate high-quality samples under high-dimensional cases, the T-SNE plots of original samples (a) and synthetic samples (b) in this case are also shown in Figure 8. Though this case is high-dimensional, it is also shown that the relative positions of synthetic samples between all the classes are similar to the relative positions of actual samples. Therefore, it also proves that the proposed method could be able to generate synthetic samples which are similar to actual samples even under a high-dimensional case.

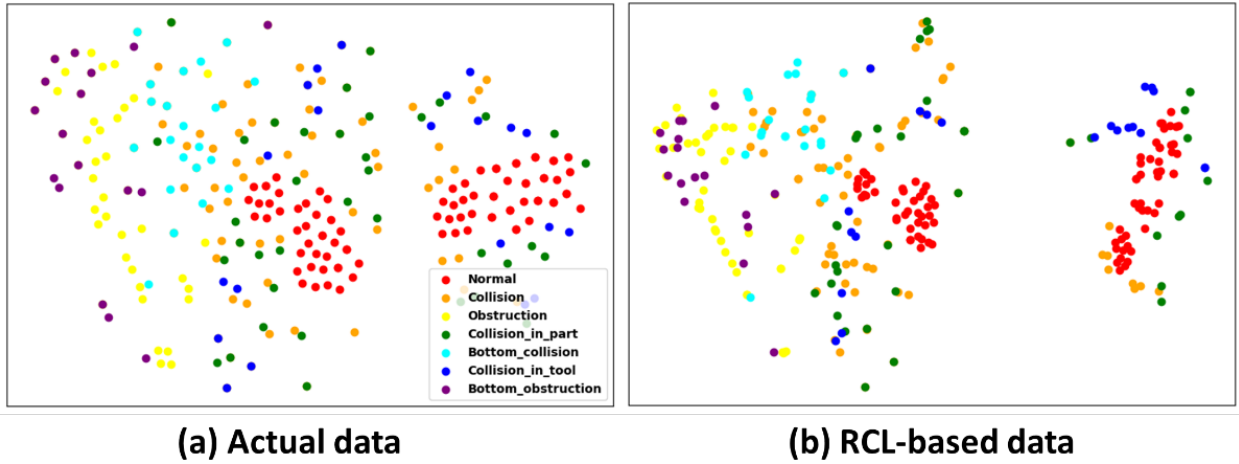

*Figure 8 The T-SNE plot of actual data (a) and RCL-based data (b).*

The precision, recall and F-score of different approaches when the number of classes increases are demonstrated in Figure 9. Besides, the specific values and standard deviations of all the metrics are shown in TABLE 2. The classification metrics of the proposed method are usually the highest, which shows that the proposed method is able to generate high-quality data. Specifically, though sometimes the proposed method has lower recalls than the baseline, the corresponding precisions and F-scores are still higher than the baseline. Especially for the last two classes, when they come in, the proposed method could always have the highest F-score. Since the last two classes have about 20 samples for training SMOTE, it also proves that 20 samples to train SMOTE is sufficient.

In addition, as the number of classes increases, all the metrics will also decrease accordingly since a greater number of classes may lead to higher difficulties of classification. However, all the metrics of the proposed method are still higher than 0.4 under the seven-class classification, which means the proposed method is still promising under a high-dimensional case. Furthermore, it is also shown in TABLE 2 that the standard deviations of the proposed method are usually lower than other approaches and comparable to the baseline. Meanwhile, the performance of the proposed method could even beat the performance of the baseline at several number of classes classification, which demonstrates that the proposed method is able to generate high-quality data without noise existing in original dataset stably. Afterwards, the case study in additive manufacturing is conducted to further validate the effectiveness of the RCL.

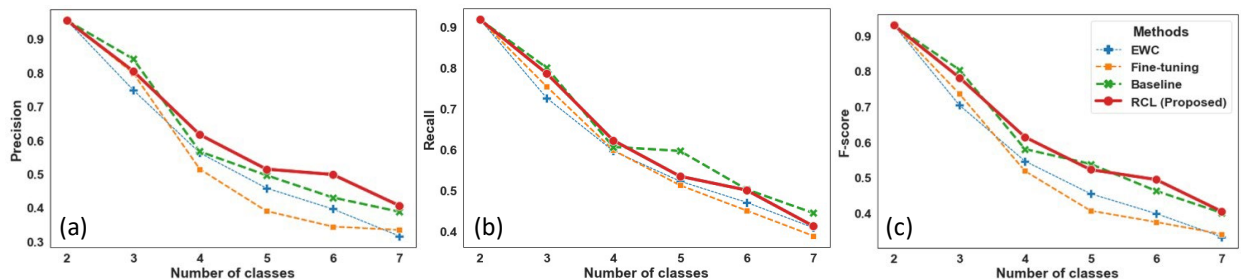

Figure 9 The precisions (a), recalls (b) and F-scores (c) of different approaches on different number of classes.

TABLE 2: Average precisions, recalls and F-scores from Task 1 to Task 6

| Method | Task 1 | | | Task 2 | | | Task 3 | | |
|---|---|---|---|---|---|---|---|---|---|
| | Precision | Recall | F-score | Precision | Recall | F-score | Precision | Recall | F-score |
| EWC | | | | 0.748 (0.039) | 0.726 (0.034) | 0.704 (0.037) | 0.562 (0.062) | 0.596 (0.045) | 0.546 (0.023) |
| Fine-tuning | 0.954 (0.022) | 0.918 (0.056) | 0.930 (0.041) | 0.798 (0.044) | 0.754 (0.049) | 0.736 (0.057) | 0.514 (0.045) | 0.598 (0.039) | 0.518 (0.043) |
| Baseline | | | | **0.840 (0.028)** | **0.800 (0.028)** | **0.804 (0.024)** | 0.566 (0.045) | 0.606 (0.026) | 0.580 (0.037) |
| **RCL (Proposed)** | | | | 0.804 (0.022) | 0.786 (0.027) | 0.782 (0.029) | **0.616 (0.033)** | **0.622 (0.030)** | **0.614 (0.033)** |

| Method | Task 4 | | | Task 5 | | | Task 6 | | |
|---|---|---|---|---|---|---|---|---|---|
| | Precision | Recall | F-score | Precision | Recall | F-score | Precision | Recall | F-score |
| EWC | 0.458 (0.023) | 0.522 (0.022) | 0.454 (0.022) | 0.396 (0.059) | 0.470 (0.017) | 0.398 (0.029) | 0.316 (0.048) | 0.408 (0.022) | 0.332 (0.027) |
| Fine-tuning | 0.390 (0.075) | 0.512 (0.027) | 0.406 (0.034) | 0.344 (0.033) | 0.450 (0.033) | 0.374 (0.027) | 0.334 (0.033) | 0.388 (0.048) | 0.340 (0.038) |
| Baseline | 0.496 (0.031) | **0.596 (0.041)** | **0.538 (0.037)** | 0.430 (0.047) | **0.502 (0.068)** | 0.462 (0.056) | 0.388 (0.024) | **0.444 (0.014)** | 0.400 (0.009) |
| **RCL (Proposed)** | **0.514 (0.035)** | 0.534 (0.023) | 0.522 (0.032) | **0.498 (0.029)** | 0.500 (0.025) | **0.494 (0.022)** | **0.406 (0.022)** | 0.412 (0.039) | **0.404 (0.032)** |

## 4.3 A real-world case study based on AM process

In this section, the sensor data collected from AM process is applied for anomaly detection while the utilization of *in-situ* accelerometers as side channels is employed for this assessment. Sec 4.3.1 elucidates the experimental setup and details the process of data collection, while Sec 4.3.2 encompasses a comprehensive presentation of the results, their interpretation, and subsequent discussion.

4.3.1 Experiment setup

In the context of this case study, a desktop fused filament fabrication (FFF)-based 3D printer, specifically Anycubic Mega-S, was employed for data collection. A vibration sensor was strategically chosen to discern motion-related alterations during the 3D printing process, a parameter intricately linked to changes during manufacturing. In alignment with the knowledge of the additive manufacturing (AM) process, the printing paths stipulated in the G-code manifest through the relative motion between the extruder and the printing bed. Consequently, for the detection of AM process alterations, one vibration sensor (MEMS accelerometer) was affixed to the extruder (see to Figure 10). The sensor is capable of recording the real-time vibrations of the extruder with approximately 1Hz sampling frequency. The data acquisition from all side channels, inclusive of the vibration sensors in this study, was executed using a Raspberry Pi 4b microcontroller.

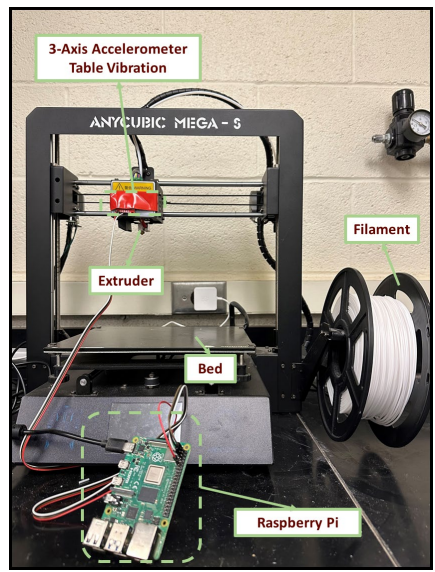

Figure 10: Experimental platform: Anycubic Mega S FFF printer.

In this study, a solid cube with an edge dimension of 2 cm served as the nominal design, incorporating the process parameters outlined in TABLE 3. Polylactic acid (PLA) filament was employed as the feedstock material. To assess the efficacy of the proposed monitoring approach, two distinct scenarios involving potential cyber-physical attacks in additive manufacturing were examined, with detailed design parameters provided in TABLE 4:

- **Case 1:** A targeted attack on layer infill rate occurred during the "slicing" stage, resulting in the modification of the infill rate.
- **Case 2:** A deliberate alteration was introduced to the design geometry during the "STL" stage, involving the insertion of a small square-shaped void.

*TABLE 3: The design parameters of nominal parts.*

| Design Parameters | Value |
|---|---|
| Printing Speed | 50 mm/s |
| Layer Thickness | 0.25 mm |
| Nozzle Temperature | 200 °C |
| Infill Rate | 20% |
| Bed Temperature | 60 °C |

*TABLE 4: The design parameters of two altered cases.*

| Design Parameters | Case 1 | Case 2 |
|---|---|---|
| Layer Design | Solid | Layer # 1~34: solid<br>**Layer # 35~80: a square hole inside** |
| Infill Rate | **25%** | 20% |
| Printing Speed | 50 mm/s | |
| Nozzle Temperature | 200 °C | |
| Bed Temperature | 60 °C | |

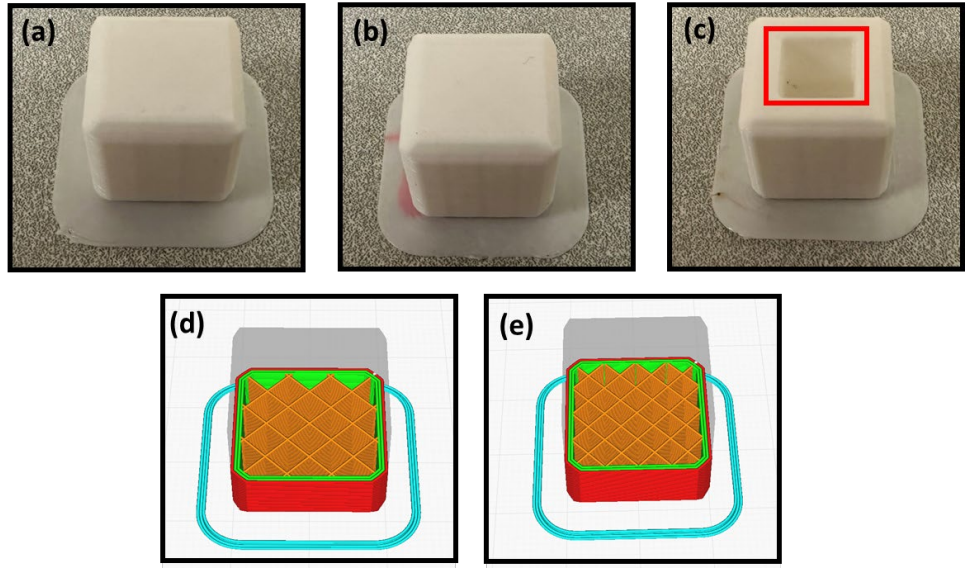

*Figure 11: Sample parts, (a) a nominal part, (b) an attacked part of Case 1, and (c) an attacked part of Case 2, (d) internal structure for nominal part, (e) internal structure for the attacked part in Case 1.*

These different cases are delineated to represent disparate defect categories. Furthermore, as illustrated in Figure 11, nominal components are present, and data acquired from these components can be construed as constituting the normal class. Consequently, the dataset is partitioned into two existing classes: normal data and data collected from case 1. Subsequently, data collected from case 2 is designated as a novel defect, indicative of a new anomaly within the memory requiring detection. As described in Sec 3.3, the class-incremental tasks are structured as follows:

- **Task 1:** Identify the defects of case 1 from the normal printing process
- **Task 2:** When the defects from case 2 come in, identify defects of both case 1 and 2 from the normal printing process

The dataset pertaining to each class encompasses two channels. Given the data is collected by time, inherent sequential information may exist. In addition, the normal and abnormal samples might be mixed up in real applications. Subsequently, window-based sampling is applied, employing a window size of 50, where the windows from normal and abnormal status could be helpful to eliminate the variation within each category.

Then each class is comprised of approximately 125 samples and five trials are conducted for each class. Consequently, the initial trial of each class, denoted as trial 1, is designated as the training set for training the proposed method. Trials 2 through 5 within each class are utilized as the testing set for performance evaluation.

In this study, a comprehensive exposition of the proposed method's significance is undertaken by scrutinizing two distinct facets. Similar to Sec 4.1, the performance evaluation is conducted through the measurement of classification metrics, specifically precisions, recall and F-score, for both Tasks 1 and 2. The effectiveness analysis is first discussed, wherein comparisons with benchmark approaches are presented. Subsequently, the flexibility analysis is illustrated, involving alterations in classifiers to ascertain the method's resilience under varied conditions.

4.3.2 Result and discussion

The parameter and benchmark setup in this case are similar as the setup in Sec. 4.1. Specifically, since the data are structured in matrix format, it is a natural idea to employ a Convolutional Neural Network (CNN) classifier. However, following some experimental trials, a Multi-Layer Perceptron (MLP) classifier may have the same performance as CNN classifier in Task 1. Therefore, to reduce the network complexity, a MLP with three fully connected layers is applied instead.

The average precisions, recalls and F-scores of all the approaches for both Task 1 and Task 2 are shown in Figure 12. The specific values of all the metrics are shown in TABLE 5 where the values in brackets are the standard deviations. All approaches, including the baseline, exhibit the same metric as 1, signifying effective accomplishment of Task 1. However, it is important to note that, such high classification metrics does not mean all the samples could be classified correctly because there are usually about 6 to 10 samples from case 1 identified as the normal class. Hence, there is still room for improvement regarding the performance of identifying the data from case 1.

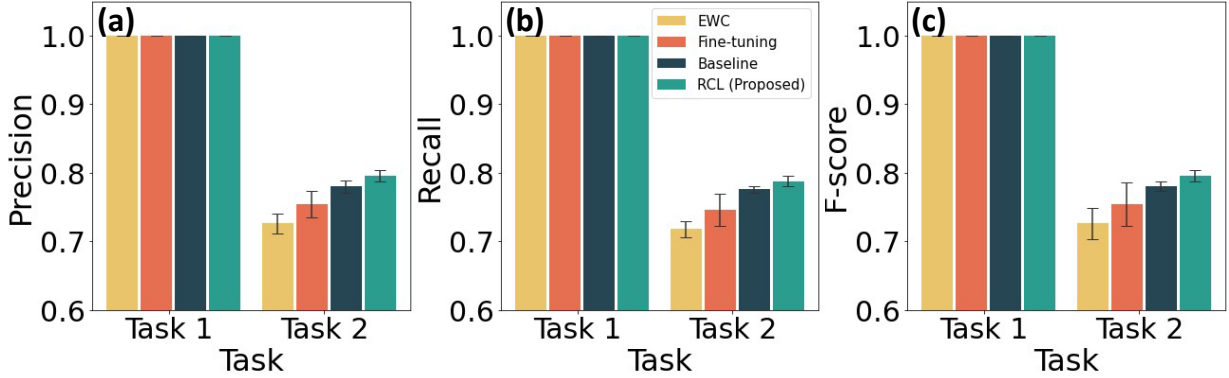

*Figure 12 The precisions (a), recalls (b) and F-scores (c) of different approaches on Task 1 and 2.*

*TABLE 5: Average precisions, recalls and F-scores of Task 1 and 2*

| Method | Task 1 | | | Task 2 | | |
| --- | --- | --- | --- | --- | --- | --- |
| | Precision | Recall | F-score | Precision | Recall | F-score |
| EWC | 1.000 (0.000) | 1.000 (0.000) | 1.000 (0.000) | 0.726 (0.014) | 0.718 (0.012) | 0.706 (0.023) |
| Fine-tuning | | | | 0.754 (0.019) | 0.746 (0.023) | 0.734 (0.032) |
| Baseline | | | | 0.780 (0.009) | 0.776 (0.005) | 0.772 (0.007) |
| **RCL (Proposed)** | | | | **0.796 (0.008)** | **0.788 (0.007)** | **0.786 (0.008)** |

In Task 2, the proposed methodology has the highest precisions, recalls, F-scores among all benchmark approaches, indicating superior performance in detecting both existing and novel defects. Notably, the performance of continual learning-based approach, usually could not exceed the baseline. However, the F-score of the proposed method is higher than the baseline, indicating that SMOTE could generate more representative data of each class, i.e., help centralizing the generated samples mitigate the bad effect of outlier data. Under such circumstances, the proposed method could further enhance performance in this class-incremental task compared with the baseline while the other approaches are not able to achieve that. In addition, it is shown that the standard deviation of the proposed method is much lower than the other approaches, showing that the proposed method could generate data stably. Hence, though the data quality may not be good, the proposed method is still able to generate more representative data to further improve the classification performance, which fully demonstrates the effectiveness of RCL framework.

Besides the outperformance of the proposed method, its flexibility is a salient aspect, as demonstrated by changing the employed classifiers. In conventional incremental learning approaches, the classifier (e.g., MLP in this study) established during Task 1, cannot change its model architecture when subsequently updated in Task 2. In contrast, the proposed method brings up much higher flexibility which allows the employment of different classifiers for these two tasks based on classifier's performance. To comprehensively illustrate this, two different classifiers are deployed for Task 2, including a MLP and a CNN classifier. According to experiments trials, the MLP classifier contains three fully connected layers while the CNN classifier contains two convolutional layers and three fully connected layers. The experiments employing different classifiers are conducted five times for Task 2. In each trial, all classifiers are trained using the same generated data. In addition, the baseline is also compared by utilizing two different classifiers in the analysis.

The average precision, recall and F-scores for Task 2 are calculated and shown in Figure 13. The specific values of all the metrics are shown in TABLE 6 where the values in brackets are the standard deviations. When using the proposed RCL method, both MLP and CNN outperform the baseline methods. Given that MLP and CNN have the same performance in Task 1, i.e., all the precisions, recalls and F-scores are 1, MLP is selected as the base model as a result of simpler model architecture. However, when Task 2 comes in, regarding training with data from the proposed RCL method, the CNN classifier has a better performance than the MLP classifier in the 3-class classification problem. This underscores the capacity of the proposed method to enhance classification performance by selecting an appropriate classifier, indicative of its heightened adaptability. Overall, the proposed method is effective in maintaining and improving defect detection performance.

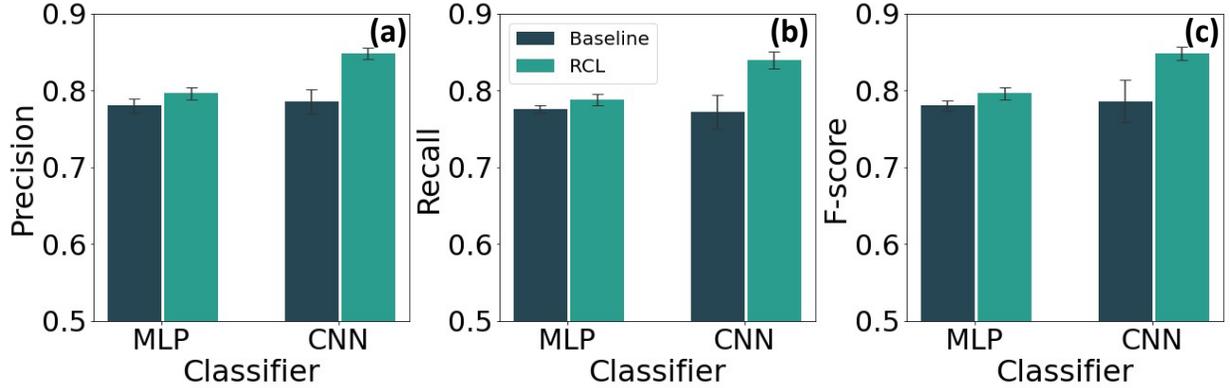

*Figure 13 The precisions (a), recalls (b) and F-scores (c) of MLP and CNN classifiers on Task 2.*

*TABLE 6: Average precisions, recalls and F-scores of Task 2 by incorporating different classifiers*

| Method | MLP | | | CNN | | |
|---|---|---|---|---|---|---|
| | Precision | Recall | F-score | Precision | Recall | F-score |
| Baseline | **0.780** (**0.009**) | **0.776** (**0.005**) | **0.772** (**0.007**) | 0.786 (0.016) | 0.772 (0.022) | 0.762 (0.028) |
| RCL (Proposed) | 0.796 (0.008) | 0.788 (0.007) | 0.786 (0.008) | **0.848** (**0.007**) | **0.840** (**0.011**) | **0.840** (**0.009**) |

# 5 Conclusions and future work

In this paper, a novel continual learning framework, denoted as the replay-based continual learning approach (RCL) using data generation, is developed to advance the capability of continual learning in classification-based anomaly detection within manufacturing system, thereby enhancing anomaly detection performance and alleviating constraints attributed to limited storage capacity. The key contributions of the proposed RCL encompass three key aspects: (1) a pseudo data generation process is proposed to surmount constraints arising from hardware limitations as the scale of data expands; (2) the integration of SMOTE helps to enhance data quality by generating more representative data compared to the original dataset, thereby potentially improving the monitoring performance; and (3) the formulation of a replay-based continual learning framework is designed to enhance flexibility by accommodating diverse model architectures between the preceding and new models based on variants in data characteristics.

The superior performance of the RCL method is validated through three real-world case studies in steel plate faults, robot execution failure and AM processes. In this empirical investigation, the proposed approach manifests the highest precision, recall, and F-scores, surpassing even the baseline, underscoring the method's efficacy in not only maintaining but also significantly enhancing monitoring performance. Furthermore, the adaptability of the proposed method is elucidated through performance evaluations under various classifiers. In the future, more experiments to explore the influence of different generators and the performance of other generative models when more data becomes available will be conducted, and more manufacturing processes with various sensing types will be utilized for evaluation.

## Notes on Contributors

**Yuxuan Li** received the B.S. degree in Statistics from Renmin University of China, Beijing, China, in 2019, and the Ph.D. degree in Industrial Engineering and Management at Oklahoma State University, Stillwater, OK, USA. He is currently an Assistant Professor in the School of Business at East China University of Science and Technology. His current research focuses on advanced data analytics in smart manufacturing and healthcare systems. He is a member of IEEE, IISE, INFORMS, and SIAM. His research has been published in IEEE Transactions on Automation Science and Engineering, IISE Transactions, Computer Methods and Programs in Biomedicine, etc.

**Tianxin Xie** received BS degrees in Agribusiness from both Oklahoma State University and China Agricultural University in 2024. She is currently pursuing a MS degree in Biostatistics at Washington University School of Medicine. Her research interest focuses on data analysis in clinical trials.

**Chenang Liu** received his double BS degrees in environmental & resource sciences and mathematics from Zhejiang University, China, in 2014; he then earned his MS degree in statistics and PhD degree in industrial and systems engineering from Virginia Tech in 2017 and 2019, respectively. He is currently an assistant professor in the School of Industrial Engineering and Management at Oklahoma State University. His research interests include data-driven analytics and machine learning-enabled modeling to advance smart manufacturing, healthcare, and service systems, as well as applied artificial intelligence for engineering applications.

**Zhangyue Shi** received a BS degree in mechanical engineering from Xi'an Jiaotong University, Xi'an, China, in 2019; he then earned his PhD degree in Industrial Engineering and Management from Oklahoma State University in 2023. He is currently a data scientist at Bayer. His current research interest includes advanced data analytics-based quality assurance in smart manufacturing and healthcare.